\documentclass{article}
\usepackage{spconf,amsmath,graphicx,hyperref,subcaption,booktabs,amssymb,microtype}
\usepackage{multirow} 
\usepackage[table, dvipsnames]{xcolor}
\usepackage{color}
\definecolor{cellcol}{gray}{.92}

\title{ Rationale-Guided Learning for Multimodal Emotion Recognition}
\name{Sujung Oh$^{1}$, Jung Uk Kim$^{2*}$, Sangmin Lee$^{3*} \thanks{*Corresponding Author}$}
\address{$^{1}$Pixel Lab, Sungkyunkwan University, South Korea\\ $^{2}$Visual AI Lab, Kyung Hee University, South Korea\\ $^{3}$Pixel Lab, Korea University, South Korea}

\makeatletter
\def\bstctlcite{\@ifnextchar[{\@bstctlcite}{\@bstctlcite[@auxout]}}
\def\@bstctlcite[#1]#2{\@bsphack
  \@for\@citeb:=#2\do{%
    \edef\@citeb{\expandafter\@firstofone\@citeb}%
    \if@filesw\immediate\write\csname #1\endcsname{\string\citation{\@citeb}}\fi}%
  \@esphack}
\makeatother
\begin{document}
%
\maketitle
\bstctlcite{BSTcontrol}

\begin{abstract}
\noindent
Multimodal emotion recognition in conversation (MERC) requires understanding complex interactions between verbal and non-verbal cues. However, most existing approaches fundamentally treat this as a direct input-output (multimodal cues-emotion labels) mapping problem, overlooking the causal reasoning that humans use when interpreting emotions. We propose rationale-guided learning (RGL), a novel framework that transforms MERC into a cognitively-inspired reasoning task. Based on dual-process theory, we decompose emotional reasoning into three facets: \textit{Intuitive} (immediate perception, \textit{System 1}), \textit{Contextual} (situational analysis, \textit{System 2}), and \textit{Integrative} (synthesis of both). We leverage a Multimodal Large Language Model (MLLM) to generate structured rationales, which are encoded as rationale banks to guide model training via aligning internal representations with human-like reasoning patterns. Our final model operates without any MLLM overheads at inference time. Experimental results show that RGL achieves state-of-the-art performance on the IEMOCAP and MELD benchmarks. Further, for interpretation, we demonstrate that the model's internal features effectively retrieve semantically correct rationales for unseen test samples, validating its rationale reasoning capabilities.

\end{abstract}
\section{INTRODUCTION}
\label{sec:intro}
Multimodal emotion recognition in conversation (MERC) aims to identify the emotional state of speakers within dialogues by leveraging text, audio, and video streams~\cite{poria2017review}. Unlike analyzing isolated clips, MERC requires a deep understanding of conversational context where emotions evolve through complex multimodal interplay. This contextual understanding is crucial for building truly empathetic interactive systems~\cite{picard1997}.

The field has progressed through several architectural shifts to better capture conversational dynamics. Initial approaches based on Recurrent Neural Networks (RNNs) \cite{majumder2019dialoguernn,hazarika2018icon} modeled dialogue sequentially, but struggled with long-range dependencies. Transformer-based models \cite{mao-etal-2021-dialoguetrm-exploring,zhang-li-2023-cross,zhao2023cfa} addressed this by leveraging self-attention to capture distant contextual cues. Subsequently, Graph Neural Networks (GNNs) \cite{shou-etal-2025-dynamic,ghosal2019dialoguegcn,hu2021mmgcn} were introduced to explicitly represent speakers and utterances as nodes, allowing more nuanced propagation of context that reflects the multi-party nature of the conversation. Alongside these architectural shifts, recent work has also emphasized robust multimodal fusion and generalization, including cross-modal knowledge distillation \cite{yun2024telme}, dynamic attention mechanisms \cite{jing2024dqformer}, and context-aware contrastive learning \cite{xie-etal-2025-dual}.

However, despite this progress, current approaches still suffer from a fundamental limitation. They treat emotion recognition as a direct mapping problem from raw inputs to emotion labels, focusing on predicting \textit{`what'} the final emotion is. This overlooks the causal reasoning humans use to understand \textit{`how'} verbal and non-verbal cues interact to convey an emotional state. For instance, wide eyes can signify fear in one context but joyful surprise in another, a nuance that rationale-free models often miss. Without modeling such rationale-driven processes, they are prone to learning superficial shortcuts from spurious correlations between input cues and emotions.

\begin{figure*}[t]
  \centering
  \includegraphics[width=\textwidth]{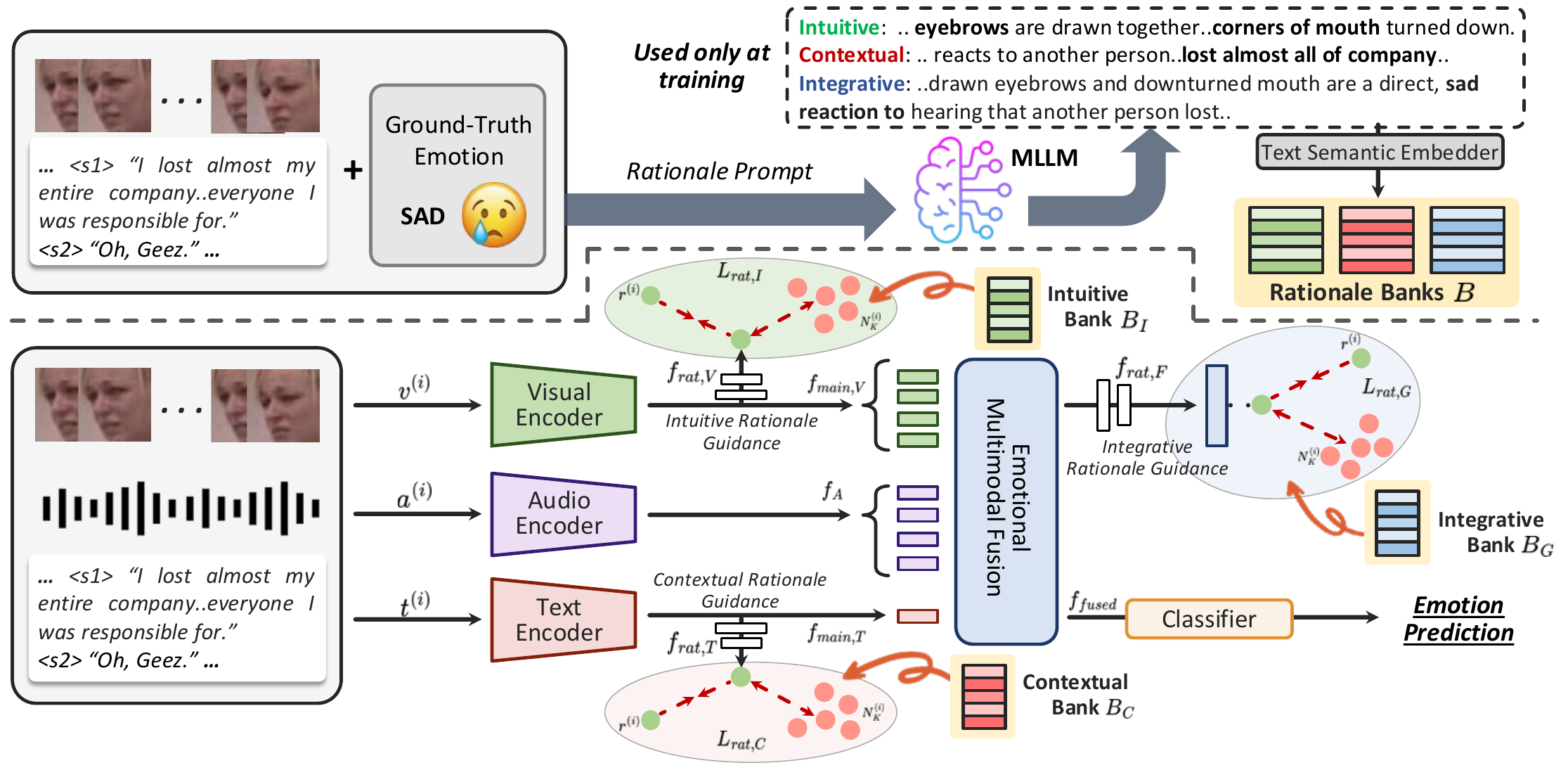} 
    \vspace{-0.7cm}
\caption{
An overview of the RGL architecture's two main stages. 
\textbf{(Top)} An MLLM generates structured rationales (\textit{Intuitive}, \textit{Contextual}, \textit{Integrative}) to construct rationale banks offline. 
\textbf{(Bottom)} A compact model is trained with contrastive losses to align its internal representations with the rationale vectors from the banks, fostering a human-like reasoning process.
}
  \label{fig:arch}
  \vspace{-0.3cm}
\end{figure*}

To address this issue, we propose Rationale-Guided Learning (RGL) for Multimodal Emotion Recognition, a novel framework that injects human-like, rationale-guided reasoning into the process using a Multimodal Large Language Model (MLLM). Crucially, the MLLM is leveraged only once during the offline training preparation step to generate rationales, allowing our final model to remain efficient without requiring any MLLM overheads at inference time. Our approach is inspired by the dual-process theory of human cognition~\cite{evans2013dualprocess}, which distinguishes between fast, automatic \textit{System 1} and slow, deliberate \textit{System 2}. Based on this theory, we decompose this reasoning into three explicit facets: \textit{Intuitive rationales} mirror the rapid perception of facial cues (\textit{System 1}), \textit{Contextual rationales} reflect the deliberate analysis of the situation (\textit{System 2}), and \textit{Integrative rationales} synthesize both (\textit{System 1 \& 2}) to form a coherent conclusion. These pre-generated rationales are transformed into intermediate supervision signals that guide our model to internalize these reasoning pathways, prioritizing causal understanding over direct input-output mappings.

Our main contributions are summarized as follows:
\vspace{-0.2cm}

\begin{itemize}
\item We propose RGL, a novel rationale-aware framework for MERC that leverages MLLM to inject human-like, rationale-guided reasoning. It enables models to learn reasoning patterns rather than superficial predictions. 
\vspace{-0.2cm}
\item We propose a three-facet rationale decomposition: \textit{Intuitive}, \textit{Contextual}, and \textit{Integrative} rationales. We utilize rationale features for intermediate guidance, enhancing the robustness without any inference overhead.
\vspace{-0.2cm}
\item Through comprehensive experiments on IEMOCAP and MELD benchmarks, we demonstrate that RGL outperforms existing state-of-the-art methods, validating the effectiveness of our rationale-aware approach.
\end{itemize}

\section{PROPOSED METHOD}
\label{sec:method}

Our proposed framework RGL, is designed to explicitly guide the reasoning process using rationales. The overall architecture is illustrated in Fig.~\ref{fig:arch}. The process consists of two primary stages: (1) Offline phase for generating three types of rationale banks (\textit{Intuitive}, \textit{Contextual}, \textit{Integrative}) by leveraging the reasoning of an MLLM, and (2) training phase for an emotion recognition model to learn rationale-guided reasoning patterns while predicting the target emotion.

\subsection{Rationale generation}
\vspace{-1mm}
\label{ssec:rationale_generation}

The cornerstone of RGL lies in structured rationale banks, generated offline by an MLLM (GPT-4o \cite{gpt4o2024systemcard}) to emulate the cognitive system from dual-process theory \cite{evans2013dualprocess}. This one-time offline process ensures our model operates without the overhead of running the MLLM at training and inference time.

We prompt the MLLM with the multimodal inputs (video frames and dialogue text) and the ground-truth emotion label for leveraging the reasoning power. The prompt is meticulously designed to guide the MLLM through a three-step analytical process, forcing it to deconstruct its reasoning into distinct, cognitively-motivated facets:
\vspace{-0.1cm}
\begin{itemize}
    \item \textbf{\textit{Intuitive rationale}} ($r_{\text{I}}$): This facet is designed to capture \textit{System 1} processing, which involves the immediate, automatic perception of evidence. The MLLM is instructed to describe only the objective facial muscle configurations (e.g., ``eyebrows are lowered and drawn together'') without using any emotional terminology.
    \vspace{-0.2cm}
    \item \textbf{\textit{Contextual rationale}} ($r_{\text{C}}$): This facet models \textit{System 2} reasoning, specifically the slower, more deliberate analysis of the surrounding situation. The MLLM identifies the specific conversational event (e.g., ``the speaker is informed their work has been shut down'') that likely triggered the emotion, which requires a deeper understanding of the dialogue's narrative.
    \vspace{-0.2cm}
    \item \textbf{\textit{Integrative rationale}} ($r_{\text{G}}$): This represents the final synthesis where the outputs of \textit{System 1} (\textit{Intuitive}) and \textit{System 2} (\textit{Contextual}) are logically connected. The MLLM formulates an explanation that justifies emotion label by combining the observed cues with the situational trigger. 
\end{itemize}
\vspace{-0.1cm}

\noindent This three-step process yields a dataset of textual descriptions for each training sample. These texts are then encoded using a pre-trained text embedder (BGE-large-en-v1.5\cite{bge_embedding}) to create dense vector representations, denoted as the rationales $\{r_{\text{I}}, r_{\text{C}}, r_{\text{G}}\}$. These rationale vectors are organized into three distinct banks, $\mathcal{B}_{\text{I}}$, $\mathcal{B}_{\text{C}}$, and $\mathcal{B}_{\text{G}}$, corresponding to the \textit{Intuitive}, \textit{Contextual}, and \textit{Integrative} facets, respectively. They serve as supervisory targets for our emotion recognition model. We refer to the combination of $\mathcal{B}_{\text{I}}$, $\mathcal{B}_{\text{C}}$, and $\mathcal{B}_{\text{G}}$ as the rationale banks $\mathcal{B}$.

\subsection{Model architecture}
\vspace{-1mm}
\label{ssec:model_architecture}
The trainable part of RGL is a compact, end-to-end network consisting of unimodal encoders and a multimodal fusion module.  

\noindent\textbf{Unimodal encoders.} 
To extract modality-specific features, our model processes visual (\textbf{V}), textual (\textbf{T}), and audio (\textbf{A}) modalities using standard pre-trained backbones: ViT-base \cite{dosovitskiy2021vit}, RoBERTa-large \cite{liu2019roberta}, and HuBERT-base \cite{hsu2021hubert}, respectively.
The visual and textual encoders are designed with a dual-head architecture to output two distinct representations: (1) the main feature $f_{\text{main,V}}$ and $f_{\text{main,T}}$ for the primary emotion prediction, (2) rationale feature $f_{\text{rat,V}}$ and $f_{\text{rat,T}}$ specifically for aligning with rationale banks.
The audio encoder outputs a single main feature, denoted as $f_{\text{A}}$. This dual-head design decouples the tasks, enabling targeted rationale alignment without interfering with the main classification objective.

\noindent\textbf{Multimodal fusion.} 
The main features from all encoders 
$\{f_{\text{main,V}}, f_{\text{main,T}}, f_{\text{A}}\}$ 
are first concatenated and then processed by a stack of Transformer encoder layers~\cite{vaswani2017attention} to model cross-modal interactions. This captures complex, cross-modal interactions through self-attention, yielding a sequence of contextually enriched hidden states $\mathbf{H} \in \mathbb{R}^{L \times D}$, where $L$ is the input sequence length and $D$ is the hidden dimension. To aggregate these sequential states into a single vector, $f_{\text{fused}}$, we employ attention pooling~\cite{lin2017structured}, which dynamically weighs the importance of each token. Finally, this vector $f_{\text{fused}}$ is projected through two task-specific heads: 
an MLP classifier for emotion prediction, and a rationale head for the rationale-guided reasoning objective. 


\subsection{Rationale-guided representation learning}
\vspace{-1mm}
\label{ssec:rationale_learning}

The core of our training is to align the model's rationale features ($f_{\text{rat,V}}$, $f_{\text{rat,T}}$, and $f_{\text{rat,F}}$) with their corresponding rationales from the pre-computed banks, $\mathcal{B}$. This alignment is achieved through a contrastive learning objective. The objective \textit{pulls} each model representation ($f^{(i)}$), referred to as the anchor, towards its corresponding target rationale ($r^{(i)}$) from the bank, which forms a \textit{positive pair}. Simultaneously, the objective \textit{pushes} the anchor away from rationales of different emotions, which form \textit{negative pairs}.

To make this process more effective, we employ a \textit{hard negative mining strategy}. For each anchor $f^{(i)}$, we construct a set of hard negatives $\mathcal{N}_K^{(i)}$ by sampling from the rationale banks $\mathcal{B}$. Specifically, we first form a candidate pool by excluding all rationales that share the same emotion label as the positive pair $r^{(i)}$. Then, from this pool, we retrieve the top-$K$ ($K=128$) most similar negative samples that have the highest cosine similarity to the anchor $f^{(i)}$. This approach forces the model to learn finer-grained distinctions between semantically close yet emotionally distinct concepts, moving beyond simple class separation.
For brevity, let $s_i^+ = \text{sim}(f^{(i)}, r^{(i)})$ denote the similarity score for the positive pair, and $s_{ik}^- = \text{sim}(f^{(i)}, r_k)$ for a negative pair $r_k \in \mathcal{N}_K^{(i)}$. The rationale loss for a sample $i$ is then defined as:
\begin{equation}
  \mathcal{L}_{\text{rat}}^{(i)} = -\log \frac{\exp(s_i^+ / \tau)}{\exp(s_i^+ / \tau) + \sum_{k=1}^{K} \exp(s_{ik}^- / \tau)},
  \label{eq:infonce_simplified}
\end{equation}
where $\text{sim}(\cdot, \cdot)$ is the cosine similarity between two vectors, $\mathcal{N}_K^{(i)}$ is the set of $K$ hard negatives for sample $i$, and $\tau$ is a temperature hyperparameter.

This alignment process is designed to mirror a cognitive reasoning pipeline inspired by the human dual-process theory.
\vspace{-2mm}

\noindent\textbf{1. Aligning visual features with \textit{Intuitive rationale}:}  
We ground the model's understanding in fast, intuitive rationale. We align the visual rationale representation $f_{\text{rat,V}}$ with the \textit{Intuitive rationale} vector $r_{\text{I}}$ from its corresponding bank $\mathcal{B}_{\text{I}}$, using the loss $\mathcal{L}_{\text{rat,I}}$.

\noindent\textbf{2. Aligning textual features with \textit{Contextual rationale}:}  
We train the model to be aware of slow, analytical reasoning of contexts from the dialogue. We align the textual rationale representation $f_{\text{rat,T}}$ with the \textit{Contextual rationale} $r_{\text{C}}$ from $\mathcal{B}_{\text{C}}$ via $\mathcal{L}_{\text{rat,C}}$.

\noindent\textbf{3. Aligning fused features with \textit{Integrative rationale}:} 
Finally, we align the fused rationale representation $f_{\text{rat,F}}$ with the \textit{Integrative rationale} vector $r_{\text{G}}$ from $\mathcal{B}_{\text{G}}$ based on $\mathcal{L}_{\text{rat,G}}$. This step guides the model to synthesize both \textit{Intuitive} and \textit{Contextual} insights, forming a coherent inference.
 This staged alignment ensures meaningful unimodal representations are learned first, providing a robust foundation for the final  synthesis.



\vspace{2mm}
\noindent
As a result, a final training objective can be formulated as:
\begin{equation}
\mathcal{L}_{\text{total}}
= \underbrace{\mathcal{L}_{\text{CE}}}_{\text{Emotion Classification}}
+ \lambda\underbrace{\bigl(\mathcal{L}_{\text{rat,I}} + \mathcal{L}_{\text{rat,C}} + \mathcal{L}_{\text{rat,G}}\bigr)}_{\text{Rationale-Guided Alignment}},
\label{eq:total_loss_priority}
\end{equation}
where $\mathcal{L}_{\text{CE}}$ represents a cross-entropy loss, each rationale loss term $\mathcal{L}_{\text{rat,X}}$ (for $X \in \{I, C, G\}$) is computed as defined in Eq.~(\ref{eq:infonce_simplified}), and $\lambda$ is a hyperparameter that balances the rationale-guided objectives, thereby training RGL to structure its embedding space in a way that mirrors a logical, human-like reasoning process.
\vspace{-0.3cm}

\begin{table}[t]
\centering
\caption{Performance comparison with existing methods on IEMOCAP and MELD datasets.}
\vspace{-3mm}
\label{tab:sota_combined}
\small
\setlength{\tabcolsep}{6pt} 
\renewcommand{\arraystretch}{1.1}
\resizebox{0.999\linewidth}{!}{\begin{tabular}{l|cc|cc}
\bottomrule
\multirow{2}{*}{\textbf{Method}} & \multicolumn{2}{c|}{\textbf{IEMOCAP}} & \multicolumn{2}{c}{\textbf{MELD}} \\ 
& \textbf{W-F1} & \textbf{Acc} & \textbf{W-F1} & \textbf{Acc} \\ 
\hline
DialogueRNN\cite{majumder2019dialoguernn} {\scriptsize (AAAI'19)} & 62.75 & 63.40 & - & - \\
DialogueTRM\cite{mao-etal-2021-dialoguetrm-exploring} {\scriptsize (EMNLP'21)} & 69.7 & 69.5 & 63.50 & 65.70 \\
MM-DFN\cite{9747397} {\scriptsize (ICASSP'22)} & 68.18 & 68.21 & 59.46 & 62.49 \\
SCFA\cite{zhao2023cfa} {\scriptsize (INTERSPEECH'23)} & 66.42 & 67.91 & 63.69 & 64.86 \\
FacialMMT\cite{zheng-etal-2023-facial} {\scriptsize (ACL'23)} & - & - & 66.58 & - \\
EASUM\cite{10484028} {\scriptsize (WACV'24)} & 69.75 & 70.10 & 65.93 & 66.70 \\
TelME\cite{yun2024telme} {\scriptsize (NAACL'24)} & 70.48 & - & 67.37 & - \\
HAUCL\cite{yi2024multimodal} {\scriptsize (ACM MM'24)} & 70.27 & 70.30 & 66.72 & 68.05
\\
BIG-FUSION\cite{wang2025bigfusion} {\scriptsize (AAAI'25)} & 72.91 & 72.64 & 67.17 & 68.24 \\
DIB-HGCN\cite{chen2025dynamic} {\scriptsize (AAAI'25)} & 72.46 & 72.58 & 66.61 & 68.01 \\
MAGTKD\cite{10.24963/ijcai.2025/905} {\scriptsize (IJCAI'25)} & 69.59 & 69.38 & 65.32 & 66.36 \\
\hline
\textbf{RGL (Ours)} & \textbf{73.68} & \textbf{73.51} & \textbf{67.43} & \textbf{68.31} \\
\toprule
\end{tabular}}
\vspace{-0.1cm}
\end{table}
\begin{table}[t!]
\centering
\caption{Ablation study of RGL's components on the IEMOCAP test set. The full model's performance is in \textbf{bold}.}
\vspace{-3mm}
\label{tab:ablation_iemocap}
\small 
\renewcommand{\arraystretch}{1.1}
\setlength{\tabcolsep}{10pt} 
\resizebox{0.85\linewidth}{!}{\begin{tabular}{l|cc}
\bottomrule
\textbf{Model Configuration} & \textbf{W-F1} & \textbf{Acc} \\\hline 
\textbf{RGL (Full Model)} & \textbf{73.68} & \textbf{73.51} \\\hline
w/o \textit{Intuitive loss} ($\mathcal{L}_{\text{rat,I}}$)
 & 72.70 & 72.52 \\
w/o \textit{Contextual loss} ($\mathcal{L}_{\text{rat,C}}$) & 68.78 & 68.70 \\
w/o \textit{Integrative loss} ($\mathcal{L}_{\text{rat,G}}$) & 72.44 & 72.34 \\
w/o $\mathcal{L}_{\text{rat,I}}$, $\mathcal{L}_{\text{rat,C}}$, $\mathcal{L}_{\text{rat,G}}$ & 68.01 & 67.71 \\\toprule
\end{tabular}}
\vspace{-0.2cm}
\end{table}

\section{EXPERIMENTS}
\label{sec:experiments}


\subsection{Datasets and implementation details}
\vspace{-1mm}
\label{ssec:implementation_details}
\noindent\textbf{Datasets.} We conduct experiments on two widely adopted datasets: IEMOCAP \cite{busso2008iemocap} and MELD \cite{poria2019meld}. IEMOCAP is a dyadic dataset for which we use six standard emotion categories:  `\textit{neutral}', `\textit{sad}', `\textit{angry}', `\textit{happy}', `\textit{excited}', and `\textit{frustrated}'. MELD is a multi-party dataset extracted from the TV show ``Friends'', containing seven emotion labels: `\textit{anger}', `\textit{disgust}', `\textit{fear}', `\textit{joy}', `\textit{neutral}', `\textit{sadness}', and `\textit{surprise}'. 

\noindent\textbf{Implementation details.} We train using the AdamW optimizer with a learning rate of $1\mathrm{e}{-5}$ and a batch size of 4. The temperature parameter in Eq. (1) is $\tau=0.07$ , and the hyperparameter in Eq. (2) is set to $\lambda=0.3$. For video streams, we follow FacialMMT~\cite{zheng-etal-2023-facial} and apply TalkNet-ASD ~\cite{10.1145/3474085.3475587} to detect the face of the active speakers based on vocal activity.

\subsection{Performance evaluation}

We evaluate performance using two standard metrics, Weighted F1 (W-F1) and accuracy (Acc), following ~\cite{shou-etal-2025-dynamic,yi2024multimodal}.

\noindent \textbf{Performance comparison.} As shown in Table~\ref{tab:sota_combined}, our RGL achieves state-of-the-art results on both datasets. The consistent improvements across two different settings, dyadic interactions on IEMOCAP and multi-party conversations in MELD, provide strong evidence for our hypothesis that explicitly supervising internal representations with structured cognitive rationales is effective.


\noindent \textbf{Ablation studies.} We also conduct ablation studies to verify the contribution of our proposed rationale designs. As shown in Table~\ref{tab:ablation_iemocap}, removing all losses simultaneously causes the most significant drop in performance, confirming their overall importance. The results reveal that $\mathcal{L}_{\text{rat,C}}$ is the most critical component, while $\mathcal{L}_{\text{rat,G}}$ and $\mathcal{L}_{\text{rat,I}}$ are also effective.

\subsection{Reasoning interpretation}
\label{ssec:qualitative}

\begin{figure}[t]
    \centering
    \includegraphics[width=0.90\columnwidth]{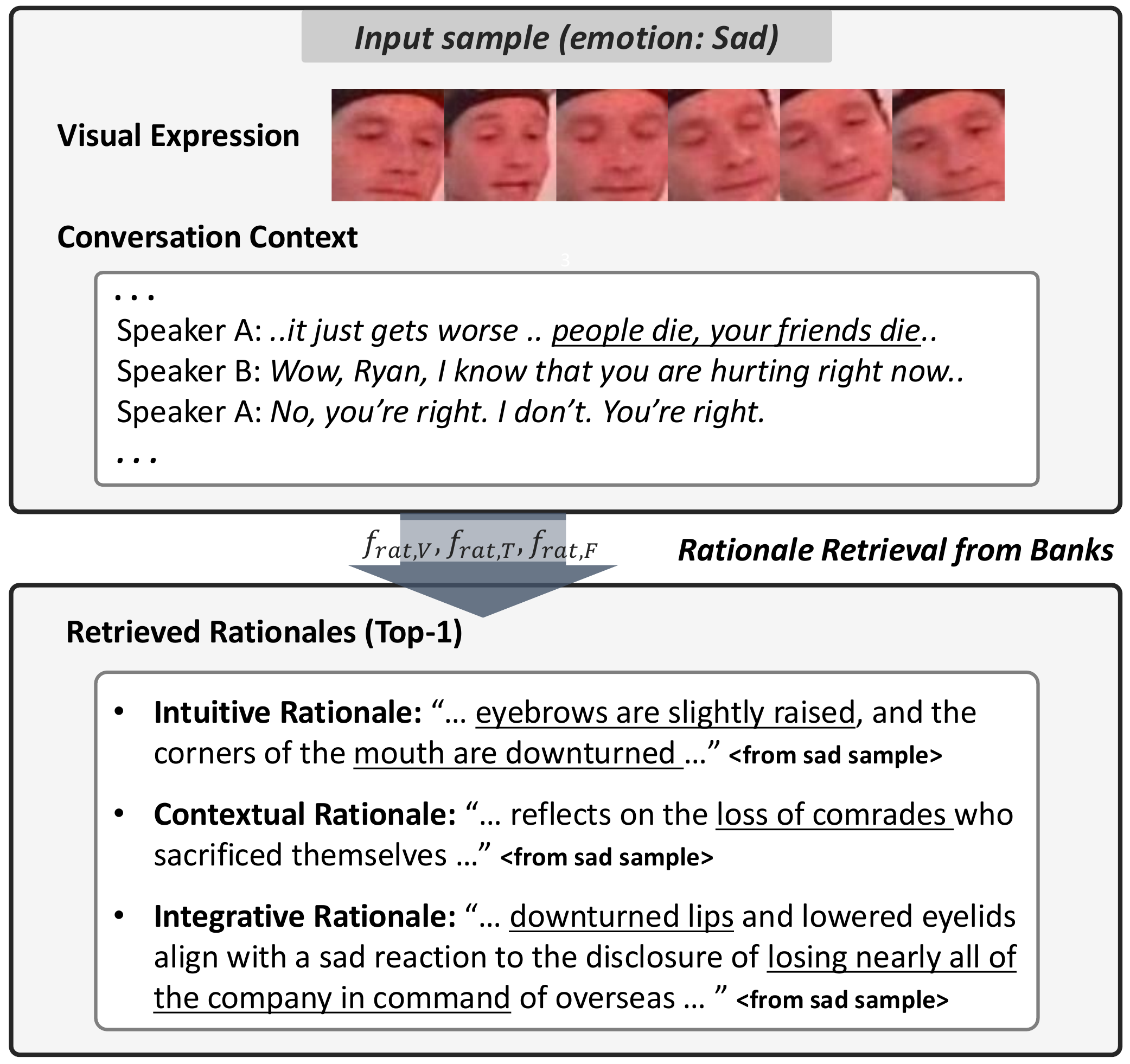}
    \vspace{-2mm}
    \caption{Rationale retrieval example on an unseen test sample.}
    \label{fig:reasoning_interpretation}
    \vspace{-5mm}
\end{figure}

To validate that RGL genuinely learns reasoning patterns, we analyze how it leverages its learned rationale banks for unseen test samples. For a given test case, we extract the model's internal rationale representations ($f_{\text{rat,V}}$, $f_{\text{rat,T}}$, $f_{\text{rat,F}}$) and use them as queries to retrieve the most similar rationales from the training rationale banks. Figure~\ref{fig:reasoning_interpretation} shows this capability on a challenging sample, where successful retrieval requires a deep semantic understanding beyond superficial cues. The model accesses an \textit{Intuitive rationale} from rationale bank that aligns with the facial expression and a \textit{Contextual rationale} relevant to the ``loss of people.'' Furthermore, the retrieved \textit{Integrative rationale} correctly synthesizes both aspects. This confirms that RGL learns a robust mapping from raw multimodal signals to a structured, semantically meaningful rationale space, proving its internal features are grounded in human-like reasoning.

\vspace{-2.5mm}
\section{CONCLUSION}
\label{sec:conclusion}

We introduce Rationale-Guided Learning (RGL) for Multimodal Emotion Recognition, a novel framework that trains a model by aligning its representations with cognitive rationales generated offline by an MLLM. Achieving state-of-the-art results on IEMOCAP and MELD, our work demonstrates that emulating cognitive reasoning is an effective approach for advancing multimodal emotion recognition.

\vspace{0.05cm}
\noindent \textbf{Acknowledgement.} This work was supported in part by the NRF grant funded by the Korea government(MSIT) (RS-2025-00563942), the IITP grant funded by the Korea government(MSIT)(IITP-2026-RS-2020-II201819, 20\%), and the IITP-ITRC grant funded by the Korea government(MSIT) (IITP-2026-RS-2023-00258649, 30\%).
\vfill\pagebreak

\label{sec:refs}


\bibliographystyle{IEEEtran} 
\bibliography{strings,refs}

@IEEEtranBSTCTL{BSTcontrol,
  CTLuse_forced_etal       = "yes",
  CTLmax_names_forced_etal = "1",
  CTLnames_show_etal       = "1",
  CTLdash_repeated_names   = "no"  
}

@article{poria2017review,
  author="Poria, Soujanya and Cambria, Erik and Bajpai, Rajiv and Hussain, Amir",
  title="A review of affective computing: From unimodal analysis to multimodal fusion",
  journal="Information Fusion",
  volume="37",
  pages="98-125",
  year="2017",
}

@Book{picard1997,
author    = "Picard, Rosalind W.",
title     = "Affective Computing",
publisher = "MIT Press",
year      = "1997",
}

@InProceedings{majumder2019dialoguernn,
author    = "Majumder, Navonil and Poria, Soujanya and Hazarika, Devamanyu and Mihalcea, Rada and Gelbukh, Alexander and Cambria, Erik",
title     = "DialogueRNN: An Attentive RNN for Emotion Detection in Conversations",
booktitle = "AAAI",
pages     = "6818-6825",
year      = "2019",
}

@InProceedings{hazarika2018icon,
author    = "Hazarika, Devamanyu and Poria, Soujanya and Mihalcea, Rada and Cambria, Erik",
title     = "ICON: Interactive Conversational Memory Network for Multimodal Emotion Detection",
booktitle = "EMNLP",
pages     = "2594-2604",
year      = "2018",
}

@inproceedings{mao-etal-2021-dialoguetrm-exploring,
author    = "Mao, Yuzhao and Liu, Guang and Wang, Xiaojie and Gao, Weiguo and Li, Xuan",
title     = "{D}ialogue{TRM}: Exploring Multi-Modal Emotional Dynamics in a Conversation",
booktitle = "Findings of EMNLP",
pages     = "2694-2704",
year      = "2021",
}

@inproceedings{zhang-li-2023-cross,
author    = "Zhang, Xiaoheng and Li, Yang",
title     = "A Cross-Modality Context Fusion and Semantic Refinement Network for Emotion Recognition in Conversation",
booktitle = "ACL-Long",
pages     = "13099-13110",
year      = "2023",
}

@InProceedings{zhao2023cfa,
author    = "Huan Zhao and Bo Li and Zixing Zhang",
title     = "Speaker-Aware Cross-Modal Fusion for Conversational Emotion Recognition",
booktitle = "INTERSPEECH",
pages     = "2718-2722",
year      = "2023",
}

@inproceedings{shou-etal-2025-dynamic,
author    = "Shou, Yuntao and Meng, Tao and Ai, Wei and Li, Keqin",
title     = "Dynamic Graph Neural {ODE} Network for Multi-modal Emotion Recognition in Conversation",
booktitle = "COLING",
pages     = "256-268",
year      = "2025",
}

@InProceedings{ghosal2019dialoguegcn,
author    = "Ghosal, Deepanway and Majumder, Navonil and Poria, Soujanya and Chhaya, Niyati and Gelbukh, Alexander",
title     = "DialogueGCN: A Graph Convolutional Neural Network for Emotion Recognition in Conversation",
booktitle = "EMNLP",
pages     = "154-164",
year      = "2019",
}

@InProceedings{hu2021mmgcn,
author    = "Hu, Jingwen and Liu, Yuchen and Zhao, Jinming and Jin, Qin",
title     = "MMGCN: Multimodal Fusion via Deep Graph Convolution Network for Emotion Recognition in Conversation",
booktitle = "ACL-Long",
pages     = "5666-5675",
year      = "2021",
}

@InProceedings{yun2024telme,
author    = "Yun, Taeyang and Lim, Hyunkuk and Lee, Jeonghwan and Song, Min",
title     = "TelME: Teacher-leading Multimodal Fusion Network for Emotion Recognition in Conversation",
booktitle = "NAACL-Long",
pages     = "82-95",
year      = "2024",
}

@InProceedings{jing2024dqformer,
author    = "Jing, Ye and Zhao, Xinpei",
title     = "DQ-Former: Querying Transformer with Dynamic Modality Priority for Cognitive-aligned Multimodal Emotion Recognition in Conversation",
booktitle = "ACM MM",
pages = "4795-4804",
year      = "2024",
}

@InProceedings{wang2025bigfusion,
  author    = "Wang, Y. and Fang, X. and Yin, H. and Li, D. and Li, G. and Xu, Q. and Xu, Y. and Zhong, S. and Xu, M.",
  title     = "BIG-FUSION: Brain-Inspired Global-Local Context Fusion Framework for Multimodal Emotion Recognition in Conversations",
  booktitle = "AAAI",
  pages     = "1574-1582",
  year      = "2025",
}

@inproceedings{xie-etal-2025-dual,
    author = "Xie, Yunhe  and
      Sun, Chengjie  and
      Cao, Ziyi  and
      Liu, Bingquan  and
      Ji, Zhenzhou  and
      Liu, Yuanchao  and
      Shan, Lili",
    title = "A Dual Contrastive Learning Framework for Enhanced Multimodal Conversational Emotion Recognition",
    booktitle = "COLING",
    pages = "4055-4065",
    year = "2025",

}

@Article{evans2013dualprocess,
author    = "Evans, Jonathan St B T and Stanovich, Keith E.",
title     = "Dual-Process Theories of Higher Cognition: Advancing the Debate",
journal   = "Perspect. Psychol. Sci.",
volume    = "8",
number    = "3",
pages     = "223-241",
year      = "2013",
}

@inproceedings{bge_embedding,
author = "Xiao, Shitao and Liu, Zheng and Zhang, Peitian and Muennighoff, Niklas and Lian, Defu and Nie, Jian-Yun",
title = "C-Pack: Packed Resources For General Chinese Embeddings",
booktitle = "ACM SIGIR",
pages = "641-649",
year = "2024",
}

@inproceedings{lin2017structured,
title     = "A structured self-attentive sentence embedding",
author    = "Lin, Zhouhan and Feng, Minwei and Santos, C{'\i}cero Nogueira dos and Yu, Mo and Xiang, Bing and Zhou, Bowen and Bengio, Yoshua",
booktitle = "ICLR",
year      = "2017",
}

@Article{gpt4o2024systemcard,
author  = "OpenAI and Ahmad, Lama and Hurst, Aaron and Lerer, Adam and Goucher, Adam P. and Perelman, Adam and Ramesh, Aditya and Clark, Aidan and Ostrow, AJ and Welihinda, Akila and et al.",
title   = "GPT-4o System Card",
journal = "arXiv",
year    = "2024",
}

@Article{liu2019roberta,
author  = "Liu, Yinhan and Ott, Myle and Goyal, Naman and Du, Jingfei and Joshi, Mandar and Chen, Danqi and Levy, Omer and Lewis, Mike and Zettlemoyer, Luke and Stoyanov, Veselin",
title   = "RoBERTa: A Robustly Optimized {BERT} Pretraining Approach",
journal = "arXiv",
year    = "2019",
}

@InProceedings{dosovitskiy2021vit,
author    = "Dosovitskiy, Alexey and Beyer, Lucas and Kolesnikov, Alexander and Weissenborn, Dirk and Zhai, Xiaohua and Unterthiner, Thomas and Dehghani, Mostafa and Minderer, Matthias and Heigold, Georg and Gelly, Sylvain and Uszkoreit, Jakob and Houlsby, Neil",
title     = "An Image is Worth 16x16 Words: Transformers for Image Recognition at Scale",
booktitle = "ICLR",
year      = "2021",
}

@InProceedings{hsu2021hubert,
author    = "Hsu, Wei-Ning and Bolte, Benjamin and Tsai, Yao-Hung Hubert and Lakhotia, Kushal and Salakhutdinov, Ruslan and Mohamed, Abdelrahman",
title     = "HuBERT: Self-Supervised Speech Representation Learning by Masked Prediction of Hidden Units",
booktitle = "ACM-TASLP",
pages     = "3451-3460",
year      = "2021",
}

@InProceedings{vaswani2017attention,
author    = "Vaswani, Ashish and Shazeer, Noam and Parmar, Niki and Uszkoreit, Jakob and Jones, Llion and Gomez, Aidan N. and Kaiser, Łukasz and Polosukhin, Illia",
title     = "Attention Is All You Need",
booktitle = "NeurIPS",
pages = "6000–6010",
year      = "2017",
}

@inproceedings{10.1145/3474085.3475587,
author = "Tao, Ruijie and Pan, Zexu and Das, Rohan Kumar and Qian, Xinyuan and Shou, Mike Zheng and Li, Haizhou",
title = "Is Someone Speaking? Exploring Long-term Temporal Features for Audio-visual Active Speaker Detection",
booktitle = "ACM MM",
pages = "3927–3935",
year = "2021",
}

@inproceedings{zheng-etal-2023-facial,
    author = "Zheng, Wenjie  and
      Yu, Jianfei  and
      Xia, Rui  and
      Wang, Shijin",
    title = "A Facial Expression-Aware Multimodal Multi-task Learning Framework for Emotion Recognition in Multi-party Conversations",
    booktitle = "ACL-Long",
    pages = "15445-15459",
    year = "2023",
}

@InProceedings{busso2008iemocap,
author    = "Busso, Carlos and Bulut, Murtaza and Lee, Chi-Chun and Kazemzadeh, Abe and Mower, Emily and Kim, Samuel and Chang, Jeannette N. and Lee, Sungbok and Narayanan, Shrikanth S.",
title     = "IEMOCAP: Interactive Emotional Dyadic Motion Capture Database",
booktitle = "LREC",
pages     = "335-339",
year      = "2008",
}

@InProceedings{poria2019meld,
author    = "Poria, Soujanya and Hazarika, Devamanyu and Majumder, Navonil and Naik, Gautam and Cambria, Erik and Mihalcea, Rada",
title     = "MELD: A Multimodal Multi-Party Dataset for Emotion Recognition in Conversations",
booktitle = "ACL",
pages     = "527-536",
year      = "2019",
}

@INPROCEEDINGS{10484028,
author    = "Hwang, Yewon and Kim, Jong-Hwan",
title     = "EASUM: Enhancing Affective State Understanding through Joint Sentiment and Emotion Modeling for Multimodal Tasks",
booktitle = "WACV",
pages     = "5668-5678",
year      = "2024",
}

@inproceedings{
yi2024multimodal,
author="Zijian Yi and Ziming Zhao and Zhishu Shen and Tiehua Zhang",
title="Multimodal Fusion via Hypergraph Autoencoder and Contrastive Learning for Emotion Recognition in Conversation",
booktitle="ACM MM",
pages = "4341–4348",
year="2024",
}

@inproceedings{chen2025dynamic,
author    = "Chen, Xuan and Shi, Weiran",
title     = "Dynamic Interactive Bimodal Hypergraph Networks for Emotion Recognition in Conversations",
booktitle = "AAAI",
pages     = "1256-1264",
year      = "2025",
}

@inproceedings{10.24963/ijcai.2025/905,
author = {Li, Jie and Ding, Shifei and Guo, Lili and Li, Xuan},
title = {Multi-modal anchor gated transformer with knowledge distillation for emotion recognition in conversation},
year = {2025},
doi = {10.24963/ijcai.2025/905},
booktitle = {IJCAI},
articleno = {905},
numpages = {9}
}

@INPROCEEDINGS{9747397,
  author="Hu, Dou and Hou, Xiaolong and Wei, Lingwei and Jiang, Lianxin and Mo, Yang",
  booktitle="ICASSP", 
  title="MM-DFN: Multimodal Dynamic Fusion Network for Emotion Recognition in Conversations", 
  pages="7037-7041",
  year="2022",
}

\end{document}